\def\year{2019}\relax
\documentclass[letterpaper]{article} 
\usepackage{aaai19}  
\usepackage{times}  
\usepackage{helvet}  
\usepackage{courier}  
\usepackage{url}  
\usepackage{graphicx}  

\usepackage{algorithm}
\usepackage{algpseudocode}
\usepackage{xcolor}
\usepackage{subcaption}
\usepackage{latexsym}
\usepackage{amsmath,amssymb} 
\usepackage{mathtools}       
\usepackage{stmaryrd}        
\usepackage{bm}              
\providecommand{\argmin}{\operatornamewithlimits{argmin}} 
\DeclareMathOperator{\Tr}{Tr}     
\DeclareMathOperator{\Var}{Var}   
\DeclareMathOperator{\Cov}{Cov}   
\providecommand{\E}{\mathbb{E}} 
\DeclarePairedDelimiterX{\inner}[2]{\langle}{\rangle}{#1, #2}
\DeclarePairedDelimiter{\norm}{\lVert}{\rVert}

\usepackage{amsthm}
\newtheorem{theorem}{Theorem}[]

\theoremstyle{definition}

\usepackage{xcolor}
\usepackage{ifthen}
\newcommand{\markupdraft}[2]{
    \ifthenelse{\equal{#1}{display}}{#2}{}
    \ifthenelse{\equal{#1}{color}}{\color{#2}}{}
}
\newcommand{\notecolored}[3][]{\markupdraft{display}{{\color{#2}\noindent[Note (#1): #3]}}}
\newcommand{\newcolored}[3][]{{\markupdraft{color}{#2}#3}
    \ifthenelse{\equal{#1}{}}{}{\markupdraft{display}{{\color{yellow!70!black}[#1]}}}}
\newcommand{\del}[2][]{{\markupdraft{display}{{\color{orange}[removed: ``#2''[#1]]}}}} 
\newcommand{\new}[2][]{\newcolored[#1]{blue}{#2}}
\newcommand{\note}[2][]{\notecolored[#1]{green}{#2}}

\newcommand{\todo}[2][]{\markupdraft{display}{{\color{red}\noindent++TODO: #2 {\color{yellow}(#1)}++}}}
\renewcommand{\markupdraft}[2]{}  

\newcommand{\shin}[1]{\note[Shinichi]{\color{magenta} #1}}

\begin{document}
%
 \title{Parameterless Stochastic Natural Gradient Method for Discrete Optimization \\
   and its Application to Hyper-Parameter Optimization for Neural Network}

   \author{K. Nishida \and H. Aguirre\\
     Shinshu University\\
     17st208e@shinshu-u.ac.jp\\
     ahernan@shinshu-u.ac.jp
     \And
     S. Saito \and S. Shirakawa\\
     Yokohama National University\\
     saito-shota-bt@ynu.ac.jp\\
     shirakawa-shinichi-bg@ynu.ac.jp
     \And
     Y. Akimoto\\
     University of Tsukuba\\
     akimoto@cs.tsukuba.ac.jp
   }

 \maketitle
\begin{abstract}
  Black box discrete optimization (BBDO) appears in wide range of engineering tasks. Evolutionary or other BBDO approaches have been applied, aiming at automating necessary tuning of system parameters, such as hyper parameter tuning of machine learning based systems when being installed for a specific task. However, automation is often jeopardized by the need of strategy parameter tuning for BBDO algorithms. An expert with the domain knowledge must undergo time-consuming strategy parameter tuning. This paper proposes a parameterless BBDO algorithm based on information geometric optimization, a recent framework for black box optimization using stochastic natural gradient. Inspired by some theoretical implications, we develop an adaptation mechanism for strategy parameters of the stochastic natural gradient method for discrete search domains. The proposed algorithm is evaluated on commonly used test problems. It is further extended to two examples of simultaneous optimization of the hyper parameters and the connection weights of deep learning models, leading to a faster optimization than the existing approaches without any effort of parameter tuning.
\end{abstract}

\section{Introduction}

The success of deep learning (DL) has changed our research field drastically. Many of previous methodologies in various fields such as natural language processing and image recognition have been replaced with DL based machine learning techniques. Many authors have developed different architectures of deep neural networks (DNN) such as VGG \cite{Simonyan2015}, ResNets \cite{He2016}, and DenseNets \cite{Huang2017} for image classification tasks, implying the importance of the design of the network configuration for promising performance of DL. However, design or selection of an appropriate DL configuration is a tedious task that requires trial-and-error based on domain specific knowledge. It prevents those who are not an expert of machine learning from applying DL based systems to their specific tasks. An \emph{automatic design of DL configuration} will enlarge applications of DL based systems with success, and hence is an important research topic.

Hyper-parameter optimization---it treats network and training configuration as hyper-parameters and optimizes them to maximize the performance after the training---is a popular approach to automatic design of DL configuration. Hyper-parameter optimization forms a Black-Box Optimization (BBO) problem as the relation between the input (hyper-parameters) and the output (outcome of training) are nearly black-box. Popular approaches to hyper-parameter optimization are evolutionary algorithms \cite{Loshchilov2016,Suganuma2017,Real2017}, Bayesian optimization approaches \cite{Snoek2012}, and reinforcement learning approaches \cite{Zoph2017}. Some other approaches couple the optimization of the hyper parameters and the connection weights \new{\cite{Srinivas2016,Shirakawa2018aaai,Pham2018icml}}. They have been applied with success in different tasks and succeeded in locating the state-o\new{f}-the-art configurations of DL.

However, full automation is jeopardized by the \emph{hyper parameters of a black box optimization algorithm}, such as the population size in evolutionary algorithms, which is critical to its performance. To distinguish the hyper parameters of DL systems and the parameters of the optimization algorithms, we call the latter as \emph{strategy parameters} in the sequel. Though it may reduce the complexity of parameter calibration, the existence of strategy parameters poses another parameter tuning, which then requires expert knowledge of the optimization algorithm. The success of covariance matrix adaptation evolution strategy (CMA-ES) \cite{Hansen:2013vt} in hyper parameter optimization for acoustic recognition tasks \cite{Watanabe2014} and in other engineering applications is partially due to its quasi-parameterless characteristic. A (quasi) \emph{parameterless BBO method} is highly desired to fully automatize the installation of DL based systems, more widely machine learning systems. It is an important direction not only for hyper-parameter optimization, but also for BBO in general.

The aim of this paper is to develop a framework of the \emph{black-box discrete optimization (BBDO)} algorithms that do not require any strategy parameter tuning by users, so that we can apply it out-of-the-box. The baseline algorithm is the Information Geometric Optimization (IGO) algorithm \cite{Ollivier2017jmlr}, which emerged from evolutionary computation as generalization of variants of CMA-ES \cite{Akimoto2012alg}, but is highly related with machine learning techniques as it is formulated as a \emph{stochastic natural gradient} \cite{Amari1998nc}. IGO is a good baseline algorithm as
generalizes not only the CMA-ES but also other known evolutionary algorithms on discrete domains that have been applied for BBDO problems such as power system controller design \cite{Folly2017}, as well as simultaneous optimization of the discrete hyper parameters and the connection weights of DL systems \cite{Shirakawa2018aaai}. Therefore, removing the need of the strategy parameter tuning for the IGO algorithm is an important research direction for BBDO including hyper parameter optimization. Though our work is highly motivated from the above mentioned hyper-parameter optimization, the algorithm developed in this paper is applicable to general BBDO problems.



We develop an adaptation mechanism for the strategy parameters of the IGO algorithm. \del{It has two strategy parameters, namely, the step-size for the parameter update and the number of Monte-Carlo samples for the natural gradient estimate. Based on theoretical investigations into the IGO algorithm in discrete search space, we fix the step-size depending only on the number of distribution parameters and adapt the number of Monte-Carlo samples }{}\new{The strategy parameters are adapted }so that the signal-to-noise ratio of the stochastic natural gradient is kept constant. The idea of the adaptation mechanism is borrowed from the so-called population size adaptation in the CMA-ES \cite{Nishida2018gecco}. We generalize the idea so that it can be applied to arbitrary discrete search domain, whereas the original one is specific for the CMA-ES in continuous domains.

As proofs of concept, we demonstrate that an instantiation of the proposed algorithm on a binary search space $\{0, 1\}^n$ improves the search efficiency over the commonly used parameter setting on standard test functions\del{ as well as max 3-SAT benchmarks}{}. We apply the proposed algorithm to the hyper parameter optimization for DL \cite{Shirakawa2018aaai} to see the efficacy of the proposed algorithm on practical examples.

\section{Information Geometric Optimization}

Information Geometric Optimization (IGO) provides a unified framework of randomized search algorithms for minimization of $f:\mathbb{X} \to \mathbb{R}$ in arbitrary search domain $\mathbb{X}$. IGO transforms the original optimization of possibly discrete variables into an maximization of a continuous variables by introducing a parametric family of probability distributions. Let $\mathcal{P} = \{P_\theta : \theta \in \Theta\}$ be a family of probability distributions $P_\theta$ on $\mathbb{X}$ parameterized by a continuous parameter $\theta \in \Theta \subseteq \mathbb{R}^n$. We assume that for all $\theta \in \Theta$, the probability measure $P_\theta$ admits the density $p_\theta$ with respect to (w.r.t.) a reference measure $\mathrm{d}x$ on $\mathbb{X}$ and the density $p_\theta$ is differentiable w.r.t.\ $\theta$. Its gradient is $\nabla p_\theta(x) = \nabla \ln(p_\theta(x)) \cdot p_\theta(x)$.

\paragraph{Stochastic Relaxation}
Let $W: \mathbb{R} \to \mathbb{R}$ be a non-increasing function. For the sake of simplicity, the readers can consider that it is the sign inversion function $x \mapsto - x$ for the moment. It transforms the objective function value to a preference (aka utility) value. Then, IGO defines the optimization problem on $\Theta$ that corresponds to the minimization of $f$ as the maximization of $J(\theta)$, where
\begin{equation}
J(\theta) := \mathbb{E}_\theta[ W(f(x)) ] = \int_{x \in \mathbb{X}} W(f(x)) p_\theta(x) \mathrm{d}x \enspace.\label{eq:sr}
\end{equation}
As long as $\mathcal{P}$ or its closure includes the Dirac delta on any $x \in \mathbb{X}$, we have $\sup_{\theta \in \Theta} J(\theta) = W(f(x^*))$, where $x^* \in \argmin_{x \in \mathbb{X}} f(x)$. This transformation turns a possibly discrete function $f$ into a differentiable function $J(\theta)$, and allows us to utilize a gradient information $\nabla J(\theta)$.

\paragraph{Natural Gradient Ascent} The function \eqref{eq:sr} is maximized by taking a gradient step $\theta \leftarrow \theta + \epsilon \mathbf{G}(\theta)^{-1} \nabla J(\theta)$, where $\epsilon$ is the \emph{step-size} (also called the \emph{learning rate}), and a matrix $\mathbf{G}(\theta)$ represents a metric of the parameter space $\Theta$, which is introduced to absorb the difference in the sensitivity of each parameter. The Fisher metric is arguably the most natural choice of the metric for $\Theta$. The resulting $\tilde \nabla = \mathbf{G}(\theta)^{-1} \nabla$ is the so called \emph{natural gradient}, where $\mathbf{G}(\theta)$ is the Fisher information matrix for $\theta$. The natural gradient achieves the invariance to coordinate system transformation on $\Theta$. In our situation, the natural gradient can be expressed as $\tilde\nabla J(\theta) = \mathbb{E}[ W(f(x)) \tilde \nabla \ln(p_\theta(x))]$.

\paragraph{Exponential Family with Expectation Parameters} An exponential family consists of probability distributions whose density is expressed as $h(x) \cdot \exp( \eta(\theta)^\mathrm{T} T(x) - \varphi(\theta) )$, where $T: \mathbb{X} \to \mathbb{R}^n$ is the sufficient statistics, $\eta: \Theta \to \mathbb{R}^n$ is the normal parameter of this family, and $\varphi(\theta)$ is the normalization factor. For the sake of simplicity, we limit our focus on the case $h(x) = 1$. If we choose the parameter $\theta$ so that $\theta = \mathbb{E}_\theta[T(x)]$, it is called the \emph{expectation parameters} of this family. Under the expectation parameters, the natural gradient of the log-likelihood is explicitly written as $\tilde \nabla \ln(p_\theta(x)) = T(x) - \theta$. The inverse Fisher information matrix is $\mathbf{G}(\theta)^{-1} = \mathbb{E}[(T(x) - \theta)(T(x) - \theta)^\mathrm{T}]$, and is typically expressed as an analytical function of $\theta$. The natural gradient step reads $\theta \leftarrow \theta + \epsilon \mathbb{E}_\theta[ W(f(x)) (T(x) - \theta)]$.

\paragraph{Stochastic Natural Gradient with Monte-Carlo} In practice we can compute neither $\mathbb{E}[ W(f(x))]$ nor $\mathbb{E}_\theta[ W(f(x)) (T(x) - \theta)]$ as $f$ is black-box. The latter needs to be approximated to perform the natural gradient ascent. This is done by employing Monte-Carlo estimation. We sample $\lambda$ points $x_i \in \mathbb{X}$ independently from $P_\theta$, which are regarded as candidate solutions of the original optimization problem, then compute $f(x_i)$ for each point. The Monte-Carlo approximation reads
\begin{equation}
  \mathbb{E}_\theta[ W(f(x)) (T(x) - \theta)]
  \approx \frac1\lambda \sum_{i=1}^{\lambda} W(f(x_i)) (T(x_i) - \theta) \enspace.\label{eq:sng}
\end{equation}
We call the right-hand side (RHS) the \emph{stochastic natural gradient}. We use the stochastic natural gradient to update $\theta$.

\subsubsection{Case of Bernoulli Distribution: cGA, UMDA, PBIL}

Given the family of Bernoulli distributions, IGO recovers several known evolutionary algorithms for optimization of binary variables: the compact genetic algorithm (cGA) \cite{Harik:1999:CGA:2221347.2221472}, the univariate marginal distribution algorithm (UMDA) \cite{Muhlenbein:1997:ERS:1326752.1326756}, and the population based incremental learning (PBIL) \cite{Baluja1995icml}. The sufficient statistics are $T(x) = x$ and the parameter $\theta$ encodes the probability of each bit to be one. That is, $\Theta = [0, 1]^n$.\footnote{
Once some components of $\theta$ become either $0$ or $1$, the distribution will never sample opposite and the parameter is fixed forever.
  To prevent this situation, the projection $\Pi = n^{-1} \vee \theta \wedge (1 - n^{-1})$ is applied after every $\theta$ update.}
In all the variants, the ranking based preference value is utilized, i.e., $W(f(x_{i:\lambda}))$ for the $i$th best candidate solution $x_{i:\lambda}$ among current $\lambda$ samples receives a pre-defined weight $w_i$, where $w_1 \geq \cdots \geq w_\lambda$.
The cGA samples only $\lambda = 2$ points and update the parameter vector $\theta$ in the direction of $x_{1} - x_{2}$ if $f(x_1) < f(x_2)$ and vice versa. It is recovered by setting the pre-defined weights as $w_1 = 2$ and $w_2 = -2$. The most commonly used value for the step size is $\epsilon = n^{-1}$. The UMDA instead set $\epsilon = 1$ while it samples multiple points $\lambda$. The pre-defined weights are  $w_1 = \cdots = w_\mu = \lambda / \mu$ and $w_{\mu+1} = \cdots w_{\lambda} = 0$, where $\mu$ denotes the number of promising points. \del{Note that the parameter update of the UMDA follows the maximum likelihood estimation of the parameter $\theta$ given promising points $x_{1:\lambda}, \dots, x_{\mu:\lambda}$. }{}The PBIL is considered as a generalization of these variants by taking $\epsilon < 1$ and $\lambda \geq 2$.


\section{Parameterless IGO Algorithm}

Our baseline algorithm is the above mentioned IGO algorithm with stochastic natural gradient. Unless explicitly mentioned, we assume an exponential family parameterized by its expectation parameters. The IGO algorithm \del{using stochastic natural gradient with Monte-Carlo sampling }{}has two strategy parameters, namely, the step size $\epsilon$ and the number $\lambda$ of Monte-Carlo samples. In practice, their suitable values depend heavily on the choice of the probability family $\mathcal{P}$ and the characteristics of the original optimization problem. To make the algorithm handy for those who are not expert of the optimization algorithm and have trouble with tuning the right values of the strategy parameters, we develop a mechanism that automatically adapts the strategy parameters during the optimization.

\subsection{Theoretical Insight into Strategy Parameters}

We provide theoretical insight into the role of the strategy parameters, by summoning up the existing analysis of related algorithms and the machinery of stochastic approximation theory, based on which we develop the adaptation mechanism for the strategy parameters.

\paragraph{Monotone Improvement of Expected Preference} An attractive property of the IGO with \emph{deterministic} natural gradient is proved in \cite{AkimotoFOGA2013}. That is, the expected preference $J(\theta) = \mathbb{E}_{\theta}[W(f(x))]$ monotonically improves over time along with the deterministic natural gradient with step size upper bounded by $1 / J(\theta)$.

\begin{theorem}[Theorem~12 of \cite{AkimotoFOGA2013}]\label{thm:qi}
  Assume that $W \circ f$ is non-negative and not almost everywhere $0$. Then, the deterministic natural gradient descent $\theta' = \theta + \epsilon \mathbb{E}_\theta[ W(f(x)) (T(x) - \theta)]$ satisfies
  \begin{equation}
    \ln \left( \frac{J(\theta')}{J(\theta)} \right) \geq  \frac{1 - \bar\epsilon}{\bar\epsilon} D_\mathrm{KL}(P_{\theta} \parallel P_{\theta'})\enspace,
  \end{equation}
  where $\bar\epsilon = \epsilon \cdot J(\theta)$. In particular, $J(\theta')\geq J(\theta)$ for $\epsilon < 1 / J(\theta)$.
\end{theorem}

An implication of this theorem is that the step size $\epsilon$ need not decrease to zero over time as long as $W \circ f$ is bounded, which can be easily achieved by taking $W: x \mapsto \exp(-x)$. Moreover, it does not need to be tuned problem by problem as we can estimate $J(\theta) = \mathbb{E}_{\theta}[W(f(x))]$ by Monte-Carlo, denoted as $\mu_W$, and set $\epsilon = \bar \epsilon / \mu_W$ for some $\bar \epsilon \in (0, 1)$. The theorem does not hold for a finite $\lambda$, and $\bar \epsilon$ may need to have a sufficiently small value. Nonetheless, the above theorem provides a useful insight into the step size setting for $\lambda$ sufficiently large.

\paragraph{Insight from Stochastic Approximation Theory}

For notation simplicity, we let $\tilde \nabla_{\infty}$ and $\tilde\nabla_{\lambda}$ denote our deterministic and stochastic natural gradient, respectively, the latter of which takes $\lambda$ samples. By introducing the time index $t$, the update is written as $\theta^{t+1} = \theta^{t} + \epsilon \tilde\nabla_\lambda^{t}$. Let us assume that $\epsilon$ is so small that the parameter vector stays in one place and the deterministic natural gradients are unchanged over $\tau > 0$ steps, i.e., $\tilde\nabla_{\infty} \approx \tilde\nabla_{\infty}^{t+s}$ for all $s \in \llbracket 0, \tau-1\rrbracket$. The parameter vector after $\tau$ steps will be approximated as
\begin{equation}
  \theta^{t + \tau} \approx \theta^{t} + (\epsilon \tau) \tilde \nabla_{\infty} + (\epsilon \tau) \frac{1}{\tau} \sum_{s=0}^{\tau-1}(\tilde\nabla_{\lambda}^{t + s} - \tilde \nabla_{\infty}) \enspace.
  \label{eq:avg}
\end{equation}
Since the true natural gradient $\tilde \nabla_{\infty}$ does not change over time, the summation term is considered as the sum of $\tau$ independent random variables with zero expectation. If we write $M = \frac{1}{\tau} \sum_{s=0}^{\tau-1}(\tilde\nabla_{\lambda}^{t + s} - \tilde \nabla_{\infty})$, its covariance matrix is
\begin{equation}
  \Cov(M) = \frac{1}{\tau} \Cov(\tilde\nabla_{\lambda}) = \frac{1}{\lambda \tau} \Cov(\tilde\nabla_{1}) \enspace.
\end{equation}
In other view, a small $\epsilon$ allows to average the stochastic natural gradient over time, and its time horizon $\tau$ will be proportional to $1/\epsilon$. Therefore, one may conclude that $\epsilon$ and $1/\lambda$ have similar effect on the accuracy of the parameter update.

\paragraph{Runtime Bound for Bernoulli Case} The first hitting time is the number of $f$-calls until the optimal solution is first hit, and is one of the most commonly used measure of the runtime of algorithms in discrete domain. \cite{Droste:2006:RAC:1152604.1152605} has shown the expected first hitting time bound on the well-known \textsc{OneMax} function and \textsc{Linear} functions. Given $\epsilon = n^{-\frac12-\delta}$ for arbitrary $\delta > 0$, the runtime of cGA on \textsc{OneMax} is $\Theta(n^\frac12 / \epsilon)$ and the lower bound of the runtime on \textsc{Linear} is $\Omega(n / \epsilon)$, where $\Theta$ and $\Omega$ are Bachmann-Landau notations. On the other hand, if \textsc{OneMax} is corrupted by an additive Gaussian noise with standard deviation $\sigma$, it is  required $\epsilon \in o((\sigma^2 n^{\frac12}\log(n) )^{-1})$ to prove the convergence of the parameter to the optimal value \cite{Friedrich2017ieee}. Interestingly, all these analyses requires $\epsilon < n^{-\frac12}$ to derive the runtime results on these simple functions. This motivates us to set $\epsilon = n^{-\frac12}$ and adapt $\lambda$.


\paragraph{Norm and Covariance of Natural Gradient}

The above theoretical results and empirical observations indicate that the step-size $\epsilon$ (more precisely, $\bar\epsilon  = \epsilon \cdot J(\theta)$) is not necessarily controlled during the optimization. It is rather different from the standard setting of stochastic gradient methods where the step-size needs to decrease in time for convergence. \del{An important characteristic that differentiates our stochastic natural gradient from the standard stochastic gradient is that the source of the randomness is from the Monte-Carlo samples drown from $P_\theta$ and resulting variance of the stochastic natural gradient depends on $\theta$. More precisely, the true natural gradient tends to the zero vector as the parameter vector approaches the Dirac delta, while the estimation covariance also converges to the zero matrix. To demonstrate it,}{}\new{To see the source of the difference,} we consider the upper bound of the squared norm of the deterministic natural gradient,
\begin{multline}
  \norm{ \mathbb{E}_\theta[ W(f(x)) (T(x) - \theta)] }^2 \\
  \leq
  \Var( W(f(x)) ) \Tr(\mathbf{G}(\theta)^{-1})  \enspace.
\end{multline}
\del{where we use the fact that the covariance matrix of the sufficient statistics $T(x)$ is the inverse Fisher information matrix $\mathbf{G}(\theta)^{-1}$. }{}On the other hand, if $W(f(x))$ is negatively or positively correlated with $T(x)$, we have
\begin{equation}
\Cov( W(f(x)) (T(x) - \theta) ) \lesseqgtr \Var( W(f(x)) ) \mathbf{G}(\theta)^{-1},
\label{eq:cov}
\end{equation}
respectively. Moreover, if $W(f(x))$ and $T(x)$ are uncorrelated, the equality holds in the above formula. Given that $W$ is bounded, $\mathbf{G}(\theta)^{-1}$ tends to zero as the distribution approaches the Dirac delta peak, implying that both the length of the natural gradient and the deviation due to the noise tend to zero at the same rate. The signal-to-noise ratio (SNR) of the stochastic natural gradient is therefore more or less constant. It is different from the natural assumption in the stochastic gradient method that its variance is uniformly bounded by a constant, e.g.\ \cite{Borkar2008book}, where the SNR of the stochastic gradient will diverge as the parameter tends to the optimum.

\subsection{Adaptation of Monte-Carlo Sample Size}

The indications of the above arguments are that both the step size and the number of Monte-Carlo samples control the estimation accuracy of the stochastic natural gradient in different manners, and one can be fixed while the other may be tuned. The theoretical result for the deterministic natural gradient update suggests to fix the step size, more precisely, fix $\bar \epsilon < 1$ and use $\epsilon = \bar \epsilon / \mu_W$ with a Monte-Carlo estimate $\mu_W$ of $J(\theta)$. \del{A reasonable range of the number of Monte-Carlo samples may depends on $\theta$, since the SNR of the stochastic natural gradient may depends. }{}Here we propose a mechanism to adapt the number of Monte-Carlo samples. We divert the idea of adaptation of the Monte-Carlo samples from the population size adaptation for CMA-ES \cite{Nishida2018gecco} and develop the adaptation mechanism for a more general case: IGO algorithm under an exponential family with expectation parameterization.

\subsubsection{Signal-to-Noise Ratio}

The main idea is to keep the signal-to-noise ratio (SNR) of the stochastic natural gradient to be constant over iterations, where we define the SNR as
\begin{equation}
  \frac{\mathbb{E}[\tilde\nabla_\lambda]^\mathrm{T} \mathbf{G}(\theta) \mathbb{E}[\tilde\nabla_\lambda]}{ \Tr( \mathbf{G}(\theta) \Cov(\tilde\nabla_\lambda) ) } \enspace.
  \label{eq:snr}
\end{equation}
The numerator is the squared norm of the expectation of the natural gradient, and the denominator is the expectation of the squared norm of the noise vector $\tilde\nabla_\lambda - \mathbb{E}[\tilde\nabla_\lambda]$, both measured w.r.t.~the Fisher metric. It will diverge to $+\infty$ as $\lambda \to \infty$ unless $\mathbb{E}[\tilde\nabla_\lambda] = \mathbf{0}$. For example, if the value of \eqref{eq:snr} is $\alpha > 0$, the norm of the expected natural gradient is equal to its standard deviation times $\alpha$.
For a technical reason we replace the numerator with the expectation of the squared norm of the stochastic natural gradient $\mathbb{E}[\tilde\nabla_\lambda^\mathrm{T} \mathbf{G}(\theta) \tilde\nabla_\lambda]$, i.e., we consider
\begin{equation}
  \frac{\mathbb{E}[\tilde\nabla_\lambda^\mathrm{T} \mathbf{G}(\theta) \tilde\nabla_\lambda] }{ \Tr( \mathbf{G}(\theta) \Cov(\tilde\nabla_\lambda) )} \enspace.
  \label{eq:snr2}
\end{equation}
Note that \del{the numerator is the sum of the numerator and the denominator of \eqref{eq:snr}. Therefore, }{}the value of \eqref{eq:snr2} is one plus the value of \eqref{eq:snr}.

\subsubsection{Averaging Over Time} As we have mentioned previously, a small step size $\epsilon$ has a similar effect as a large sample size $\lambda$. However, the effect of $\epsilon$ will not appear in the above defined SNR \eqref{eq:snr2}. To take the effect of the step size into account, we consider the SNR of the stochastic natural gradient averaged over time. Let $\beta$ be the inverse time horizon of the averaging. We introduce an accumulation of the stochastic natural gradient, that is,
\begin{equation}
  \bm{s}^{(t+1)} = (1 - \beta) \bm{s}^{(t)} + \sqrt{\beta(2 - \beta)} \frac{\mathbf{G}(\theta^{(t)})^{\frac12} \tilde \nabla_{\lambda}^{(t)}}{ \Tr( \mathbf{G}(\theta) \Cov(\tilde\nabla_\lambda) )^\frac12} \enspace,\label{eq:s}
\end{equation}
where $\bm{s}^{(0)} = \bm{0}$. Its squared Euclidean norm is compared with $\gamma^{(t+1)} = (1 - \beta)^2 \gamma^{(t)} + \beta(2 - \beta)$, where $\gamma^{(0)} = 0$. If $\beta = 1$, the expectation of $\norm{\bm{s}^{(t+1)}}^2 / \gamma^{(t+1)}$ is simply \eqref{eq:snr2}.

To see the effect of $\beta$, we consider the same situation as in \eqref{eq:avg}, i.e., $\epsilon$ is small enough that $\theta$ stays in one place. Then, the second term on the RHS of \eqref{eq:s} can be regarded as i.i.d.~samples, and $\bm{s}$ tends to $\sqrt{(2-\beta) / \beta} \mathbf{G}(\theta)^{\frac12} \mathbb{E}[\tilde \nabla_{\lambda}] / \Tr( \mathbf{G}(\theta) \Cov(\tilde\nabla_\lambda) )^\frac12$ with noise having the identity covariance matrix. In other words, the signal is enhanced by factor $\sqrt{(2-\beta) / \beta}$ while the noise strength is not affected by $\beta$. As a result, the expectation of $\norm{\bm{s}^{(t+1)}}^2 / \gamma^{(t+1)}$ approaches $(2 - \beta) / \beta$ times the SNR \eqref{eq:snr} plus one. The idea is to set $\beta = \epsilon$ so that the effect of a small $\epsilon$ will be taken into consideration.

\subsubsection{Uncorrelated Scenario} We still need to estimate in \eqref{eq:s} the covariance matrix of the stochastic natural gradient. One can estimate it by bootstrap. However, it is not reliable particularly when the number of samples are smaller than the number of parameters, $\lambda < n$, which is the usual situation. Therefore, we will replace it with a reasonable alternative value.

We consider the case where $f(x)$ and $T(x)$ are uncorrelated. In this situation, the expectation of the stochastic natural gradient is zero, and its covariance matrix is $\Cov(\tilde\nabla_\lambda) = \Var( W(f(x)) ) \mathbf{G}(\theta)^{-1} / \lambda$ as deduced from \eqref{eq:cov}. Substituting it, we obtain
\begin{equation}
  \Tr( \mathbf{G}(\theta) \Cov(\tilde\nabla_\lambda) )^\frac12 = \sqrt{(n / \lambda) \Var( W(f(x))) } \enspace.
\end{equation}
The variance $\Var( W(f(x))) $ of the preference is estimated from the Monte Carlo samples.

\subsubsection{Adaptation Mechanism}

Let $\alpha \geq 1$ be the target SNR level. We aim at keeping $\norm{\bm{s}^{(t+1)}}^2 / \gamma^{(t+1)} \approx \alpha$ by adapting $\lambda$. Let $\lambda_r$ be a real-valued intermediate sample size. Every iteration the sample size is updated by
\begin{equation}
  \lambda_r \leftarrow \lambda_r \exp\left( \beta \left( \gamma^{(t+1)} - \norm{\bm{s}^{(t+1)}}^2 / \alpha  \right) \right) \enspace,
\end{equation}
and clipped within $[\lambda_{\min}, \lambda_{\max}]$, where they are the predefined minimum and maximum sample sizes.
The sample size is then updated as $\lambda = \text{round}(\lambda_r)$.

\subsection{Algorithm Summary and Default Parameter}

The proposed stochastic natural gradient descent with adaptive Monte-Carlo sample size, called \emph{parameterless IGO}, is summarized as follows.
We introduce two minor changes.
One is that the average preference value is subtracted from the preference value. It is a well-known technique to reduce the estimation variance of gradient while the expectation is unchanged, see e.g., \cite{Evans2000book}.
The other is that we take a projection $\Pi$ so that $\theta$ stays inside $\Theta$.
\begin{algorithm}\caption{Parameterless IGO}\label{algo:pigo}
\begin{algorithmic}[1]
  \Require{$\theta$} \Comment{initial distribution parameter}
  \Require{$\epsilon  = \beta = n^{-\frac12}$, $\alpha = 1.5$, $\lambda_{\min} =2$, $\lambda_{\max} = n$} 
  \State $\lambda = \lambda_r = \lambda_{\min}$, $\gamma = 0$, $\bm{s} = \bm{0}$
  \Repeat
  \State sample $\lambda$ points $x_i$ independently from $P_\theta$
  \State evaluate $W(f(x_i))$ for all $i = 1, \dots, \lambda$
  \State $\mu_W = \frac1\lambda\sum_{i=1}^\lambda W(f(x_i))$
  \State $\sigma_W^2 = \frac{1}{\lambda}\sum_{i=1}^\lambda (W(f(x_i)) - \mu_W)^2$
  \If{$\sigma_W^2 = 0$}
  \State skip the following
  \EndIf
  \State $\tilde\nabla_\lambda = \frac1\lambda \sum_{i=1}^{\lambda} (W(f(x_i)) - \mu_W) (T(x_i) - \theta)$
  \State $\theta \gets \Pi(\theta + (\epsilon / \mu_W) \tilde\nabla_\lambda)$  
  \State $\bm{s} \gets (1 - \beta) \bm{s} + \sqrt{\beta(2 - \beta) \lambda / (n \sigma_W^2)} \mathbf{G}(\theta)^{\frac12} \tilde \nabla_{\lambda}$
  \State $\gamma \gets (1 - \beta)^2 \gamma + \beta(2 - \beta)$
  \State $\lambda_r \gets \lambda_{\min} \vee \lambda_r \exp( \beta ( \gamma - \norm{\bm{s}}^2 / \alpha )) \wedge \lambda_{\max}$
  \State\label{l:lam} $\lambda = \text{round}(\lambda_r)$  
  \Until{termination condition are met}
\end{algorithmic}
\end{algorithm}

The necessary input to this algorithm is the initial distribution parameter $\theta$. This parameter should be chosen so that the least prior information is assumed, unless one has a good candidate for the optimal $\theta$ for the specific problem. \del{For example, if we consider the Bernoulli distribution, where $T(x) = x$ and $\theta \in [n^{-1}, 1-n^{-1}]^{n}$, $\theta = (0.5, \dots, 0.5)$ is the candidate of the initial distribution parameter since all the state $x \in \{0, 1\}^n$ have the same probability.}{}

The algorithm has five strategy parameters: step size $\epsilon$, inverse time horizon $\beta$ of averaging, threshold $\alpha$ for SNR, minimum and maximum sample sizes $\lambda_{\min}$ and $\lambda_{\max}$. The algorithm is quasi-parameter-less, in the sense that all the strategy parameters have their default values depending solely on the number of parameters $n$ and they are not meant to be tuned problem by problem. The threshold $\alpha = 1.5$ is determined based on our preliminary experiments. The higher $\alpha$ is, the higher the sample size is kept. As we mentioned above, $\beta = \epsilon$ is required to enhance the signal averaged over time. The step size can be even one as long as $\lambda$ is large enough, but it will leads to undesired drift of $\theta$ when $\lambda$ is not sufficiently large. To stabilize the parameter update, we select $\epsilon = n^{-\frac12}$, motivated by the runtime analysis of cGA mentioned above. The minimum sample size $\lambda_{\min} = 2$ is a requisite for computation of $\sigma_W^2$, and the maximum number $\lambda_{\max}$ is more like an option. This can be $\lambda_{\max} = \infty$, since the adaptation mechanism will not let $\lambda$ diverge.

\subsubsection{Adaptive Step-Size}

It is often the case in black box scenario that the $f$-call is the computational bottleneck and one wants to parallelize $f$ evaluations. For this purpose, a large $\lambda$ is preferred to a small $\epsilon$. \del{Therefore, we suggest to adapt $\lambda$, while keeping $\epsilon$ constant in general. }{}However, there are cases that we want to keep $\lambda$ constant while decreasing $\epsilon$ instead. \cite{Shirakawa2018aaai} is one such case, where the weights of the neural network is updated every iteration of the algorithm, and we would like to have more iterations with a small $\lambda$, rather than less iterations with a large $\lambda$.
The same technique can be used to adapt the step-size $\epsilon$ while keeping $\lambda = \lambda_{\min}$. Line~\ref{l:lam} is replaced with $\lambda = \lambda_{\min}$ and $\epsilon = \beta = n^{-\frac12} / (\lambda_r / \lambda_{\min})$. That is, \del{$\lambda$ is fixed to $\lambda_{\min}$, while }{}$\epsilon$ and $\beta$ are decreased from the default value $n^{-\frac12}$ by factor $(\lambda_r / \lambda_{\min})^{-1}$. Then, $\epsilon$ is adapted within $[(\lambda_{\min} / \lambda_{\max}) n^{-\frac12}, n^{-\frac12}]$.

\subsubsection{Instantiation with Bernoulli: Parameterless PBIL}

We consider an instantiation of the parameterless IGO algorithm with Bernoulli distributions. The sufficient statistics are $T(x) = x$, and
the Fisher information matrix $\mathbf{G}(\theta)$ is the diagonal matrix whose $k$th diagonal element is $(\theta_k(1-\theta_k))^{-1}$. The initial parameter should be $\theta = (0.5, \dots, 0.5)$ unless one has a prior knowledge into a good region. As mentioned in the previous section, the IGO framework recovers several known evolutionary algorithms such as cGA, UMDA, and PBIL. Following the existing studies, we employ the ranking based preference function, i.e., $i$th best solution $x_{i:\lambda}$ among current $\lambda$ solutions receives $W(f(x_{i:\lambda})) = w_i$, where $w_i$ are pre-defined weights. To generalize cGA, which is the most commonly used variant among the above, we set $\mu = \lceil\lambda/4\rceil$ and $w_i = 2 \lambda/\mu$ for $i \leq \mu$, $w_i = \lambda / \mu$ for $\mu < i \leq \lambda-\mu$, and $w_i = 0$ for $i > \lambda - \mu$. If there are ties, the weights are averaged and distributed equally to the ties. Then, line~6 of the algorithm reads $\mu_W = 1$. If there is no tie, we also have $\sigma_W^2 = 2 (\lambda / \mu) \approx 8$. We remark that $\epsilon = 1/n$ is the most commonly used setting for $\lambda = 2$ (cGA), whereas we use $\epsilon = n^{-\frac12}$ as the default value for our proposed approach.

\section{Benchmark Results}
\newcommand{\pbill}{PBIL-$\lambda$}
\newcommand{\pbile}{PBIL-$\epsilon$}

The proposed \emph{Parameterless PBIL} has been compared with cGA. The parameterless PBIL with $\lambda$ and $\epsilon$ adaptation are denoted as \pbill\ and \pbile, respectively.

We consider two commonly used test functions, \textsc{OneMax} and \textsc{LeadingOnes}. They are defined as $\textsc{OneMax}(x) = n - \sum_{k=1}^{n} x_k$ and $\textsc{LeadingOnes}(x) = n - \sum_{k=1}^{n} \prod_{j=1}^{k}x_j$. \del{Both have the optimal value $0$, achieved at $x = (1, \dots, 1)$. }{}We run ten independent trials for each algorithm on each function with $n \in [10, 3000]$. All the algorithms are computationally cheap and usually the simulation for the objective is the bottleneck. We report the number of $f$-calls until they find the optimum as the runtime measure.

\begin{figure}
	\centering
	\includegraphics[width=\hsize]{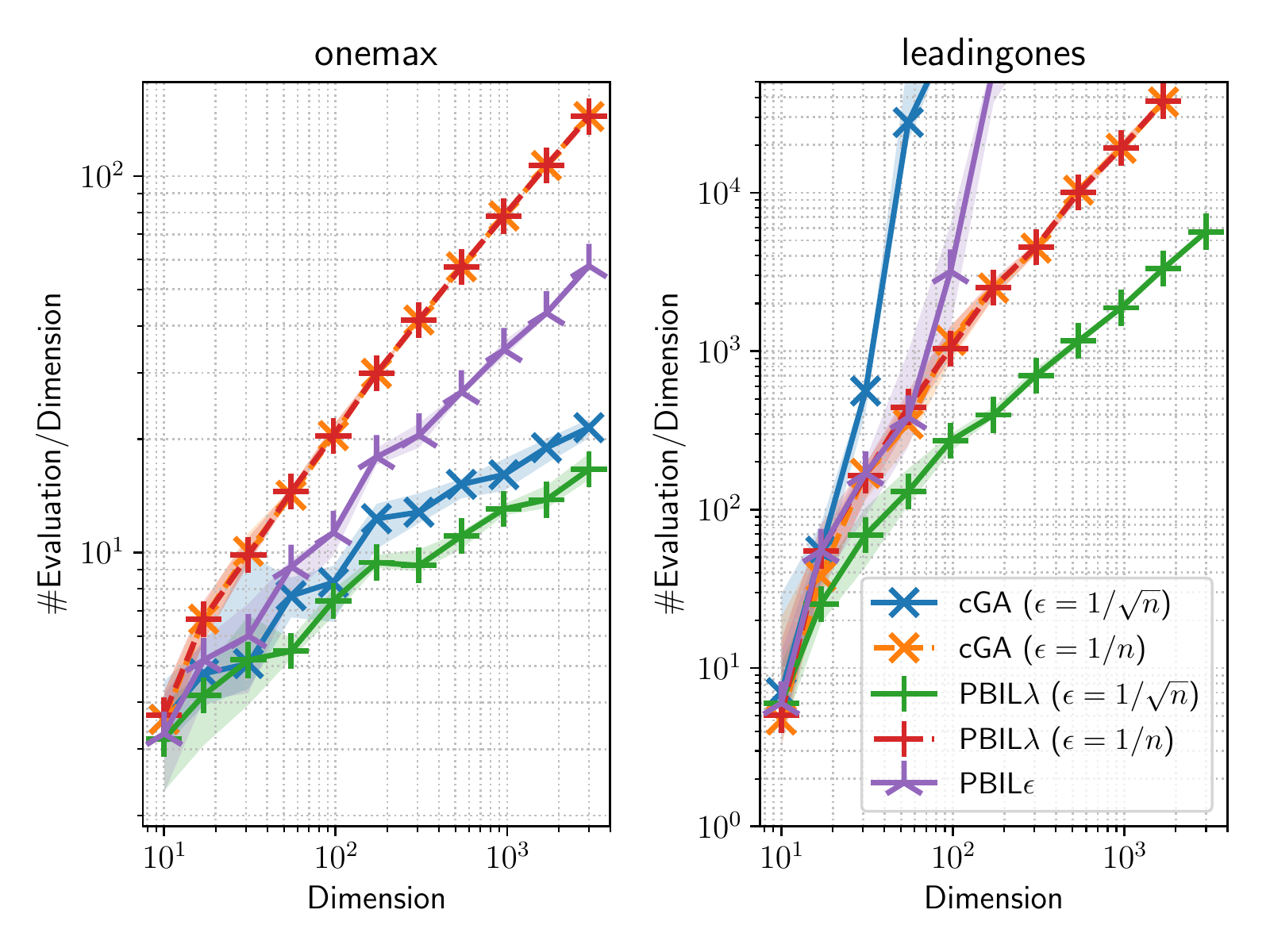}
	\caption{Parameterless PBIL v.s.\ cGA.  Median (solid symbol) and inter-quartile range (transparent region) of the first hitting time are displayed.}
	\label{fig:sp1}
\end{figure}

The indications of Figure~\ref{fig:sp1} is summarized as follows. The optimal $\epsilon$ of cGA depends on problems. On \textsc{OneMax} $\epsilon = n^{-\frac12}$ resulted in significantly faster convergence, whereas it was too large for \textsc{LeadingOnes} and cGA failed to locate the optimum. With $\epsilon = n^{-1}$, the proposed method performed very similarly to cGA, as it kept $\lambda = 2$. With $\epsilon = n^{-\frac12}$ (our proposed setting), the proposed method succeeded to locate the optimum with fewer number of $f$-calls than cGA with $\epsilon = n^{-\frac12}$ on \textsc{OneMax} and cGA with $\epsilon = n^{-1}$ on \textsc{LeadingOnes}. During the experiments, $\lambda$ was typically increased to around $16$ for $n = 1000$, interestingly, on both functions. However, $\lambda$ dynamics were different. On \textsc{OneMax} it tended to keep $\lambda$ constant, whereas on \textsc{LeadingOnes} $\lambda$ was increasing as the distribution parameters approach the optimum values. The proposed method does, not only alleviate a painful parameter tuning, but also take advantage of the dynamic strategy parameter to speedup the search. \pbile\ tended to be slower than \pbill\ on both functions, and scaled even worse than cGA with $\epsilon = n^{-1}$ on \textsc{LeadingOnes}, though it adapted $\epsilon < n^{-1}$. The defect of \pbile\ is that $\theta$ randomly drifts away from the initial value before $\epsilon$ is well adapted, and it is difficult to get back as $\epsilon$ is small. Robustness of \pbill\ against $\alpha$ variation was investigated in Appendix (supplementary material).

\del{The algorithms were also compared on Max 3 SAT problem instances, taken from SATLIB\footnote{http://www.cs.ubc.ca/~hoos/SATLIB/benchm.html}.
The median performances of cGA ($\epsilon = n^{-1}$), \pbill, and \pbile\ were similar, while the inter-quartile range of the performance of cGA was significantly greater than those of \pbill\ and \pbile. The reason may be because the adequate $\epsilon$ values for cGA are different for different problem instances. See Appendix for details.}{}

\section{Hyper-Parameter Optimization}

The simultaneous optimization of hyper parameters and connection weights for DL has been developed in \cite{Shirakawa2018aaai}. Let $\phi(W, M)$ be a DNN with connection weight $W$ and hyper parameter $M \in \mathcal{M}$. The latter, $M$, encodes, for example, the type of activation function of each unit, and we limit our focus to the case $\mathcal{M} = \{0, 1\}^n$. Let $\mathcal{L}(W, M)$ be the true loss of the model. This loss is typically differentiable w.r.t.\ $W$, while obviously it is not w.r.t.\ $M$. Introducing $P_\theta$ on $\mathcal{M}$ and applying the same idea as IGO, we obtain the differentiable objective function, $\mathcal{G}(W, \theta) = \E_{\theta}[\mathcal{L}(W, M)]$. Then, we take the stochastic gradient w.r.t.~$W$ and stochastic natural gradient w.r.t.~$\theta$ in the same way as in IGO. Given the Bernoulli distribution $P_\theta$, they read
\begin{align}
  \nabla_W \mathcal{G}(W, \theta) &\approx \textstyle \frac{1}{\lambda} \sum_{i=1}^{\lambda} \nabla_W \mathcal{L}(W, M_i; \mathcal{Z}) \label{eq:nggw}\\
  \tilde\nabla_\theta \mathcal{G}(W, \theta) &\approx  \textstyle\frac{1}{\lambda} \sum_{i=1}^{\lambda} \mathcal{L}(W, M_i; \mathcal{Z})  (M_i - \theta) \enspace,\label{eq:nggt}
\end{align}
where $\mathcal{Z}$ represents a mini-batch of training data\del{ with size $N$}{}, and $\mathcal{L}(\cdot; \mathcal{Z})$ is the average loss associated to $\mathcal{Z}$. The loss in \eqref{eq:nggt} is replaced with the ranking based preference as done in PBIL. Every iteration, $W$ and $\theta$ are updated using the above stochastic (natural) gradient using mini-batch $\mathcal{Z}$ and Monte-Carlo samples $(M_i)_{i=1}^{\lambda}$. In the reference, $\lambda = 2$ and \del{the step-size for $\theta$ update is }{}$\epsilon = n^{-1}$, recovering cGA \new{with the most commonly used setting}. From the viewpoint of cGA, the objective function, $f(M) = \mathcal{L}(W, M; \mathcal{Z})$, is dynamic since $W$ changes over iterations and stochastic since a different $\mathcal{Z}$ comes at each iteration.

We can seamlessly incorporate our proposed parameter adaptation mechanisms into the above framework as the $\theta$ update is regarded as cGA. It is expected to speedup the optimization of $\theta$ without any effort for parameter tuning. Moreover, since the objective is essentially regarded as noisy, a dynamic change of strategy parameters is encouraged in theory \cite{Friedrich2017ieee}. We apply \pbill\ and \pbile\ as alternatives to cGA, however, the former has a systematic disadvantage over cGA and \pbile. Since $W$ is updated as often as $\theta$ is done in the original setting, \pbill\ will update $W$ less often (as $\lambda$ will be greater), leading to a slow convergence of $W$. To make \pbill\ competitive, we update $W$ after every process of $\phi$ for \pbill. \del{, i.e., $W$ in \eqref{eq:nggt} are different for different $M_i$.}{}\new{For each $M_i$, we draw a new mini-batch $\mathcal{Z}_i$ and compute the loss $\mathcal{L}(W, M_i; \mathcal{Z}_i)$. Then $W_i$ is updated using $\nabla_W\mathcal{L}(W, M_i;\mathcal{Z})$.} As $M_i$ evaluated later tends to have better loss (since $W$ is updated), it will introduce a bias in the stochastic natural gradient, yet it is more promising than just updating $W$ after every $\lambda$ processes of $\phi$\del{ (i.g., $\lambda$ GPU processes)}{}. The performance differences between cGA and \pbile\ are purely due to the adaptation of $\epsilon$, while the differences between \pbile\ and \pbill\ may be partly due to the above modification. See Appendix for details.

We performed two experiments; experiment I: selection of layers, and experiment II: selection of activation functions. For fair comparisons, we followed the experimental setup in \cite{Shirakawa2018aaai}, except the above mentioned modification for \pbill. In both experiments, a feed-forward network with fully connected hidden layers were trained on MNIST handwritten digits dataset. \new{The mini-batch size for training is $64$ for all cases. }The weight $W$ is updated with Nesterov's momentum \cite{Sutskever2013} of 0.9, and a weight decay of $10^{-4}$ is employed. The step-size for $W$ update is divided by 10 at 1/2nd and 3/4th of the maximum \del{epochs}{}\new{iterations} \shin{In AAAI '18, I use the epoch for decreasing timing but in this paper we use iteration.}\todo{shouldn't it be the maximum iterations? The numbers of iterations for tanh (or ReLU) and cGA given maximum epochs are different, as cGA spend half of mini-batch at one iteration.}\cite{He2016,Huang2017}. \del{The mini-batch size is 128 for fixed $M$ cases and for \pbill. For cGA and \pbile, it is halved and two network processes for one iteration are done at once in GPU, in order to update $W$ after every GPU process. This way, the computational time for all the variants are competitive as the number of GPU processes is the same for a fixed iterations. }{}The cross entropy error with softmax activation is used as the loss $\mathcal{L}$. When we predict the test data, the hyper parameter $M = (m_1, \dots, m_n)$ is fixed such that $m_k = 1$ if $\theta_k \geq 0.5$, otherwise $0$. See \del{\cite{Shirakawa2018aaai}}{}\new{the reference} for further details. We conducted 30 trials for each setting. They are implemented in Python using \texttt{Chainer 4.4.0}. 

\paragraph{(I) Selection of Layers}
A DNN with 32 fully connected hidden layers with 128 units and ReLU \cite{Nair2010} activation was trained. A binary vector $M \in \{0, 1\}^n$ of dimension $n = 31$ determines how successive layers are connected. Let $X_k$ be the input to the $k$th \new{hidden} layer and $H_k$ is the operation of $k$th layer. The input to the $k+2$nd layer $X_{k+2}$ (for $k = 1, \dots, n$, where $X_{1}$ is equal to the input to the network and $X_{33}$ is regarded as the input to the output layer) is modeled as $X_{k+2} = X_{k+1} + m_k \cdot H_{k+1}(X_{k+1})$, i.e., the operation of the $k+1$st layer is skipped if $m_{k} = 0$.\footnote{\cite{Shirakawa2018aaai} modeled the process as $X_{k+2} = (1 - m_k) \cdot X_{k+1} + m_k\cdot H_{k+1}(X_{k+1})$. However, our preliminary experiment found our setting is more stable due to the skip connection from the previous layer. We also adopt the gradient clipping with the norm of $2$ to prevent too long gradient step.} The number of effective hidden layers are then $1 + \sum_{k=1}^n m_k$. Note the given model is overly deep.\note{the best number of layers was ten, whose test error is 1.778}

\begin{figure}[t]
  \centering%
  \begin{subfigure}{0.5\hsize}%
    \centering%
    \includegraphics[width=\hsize]{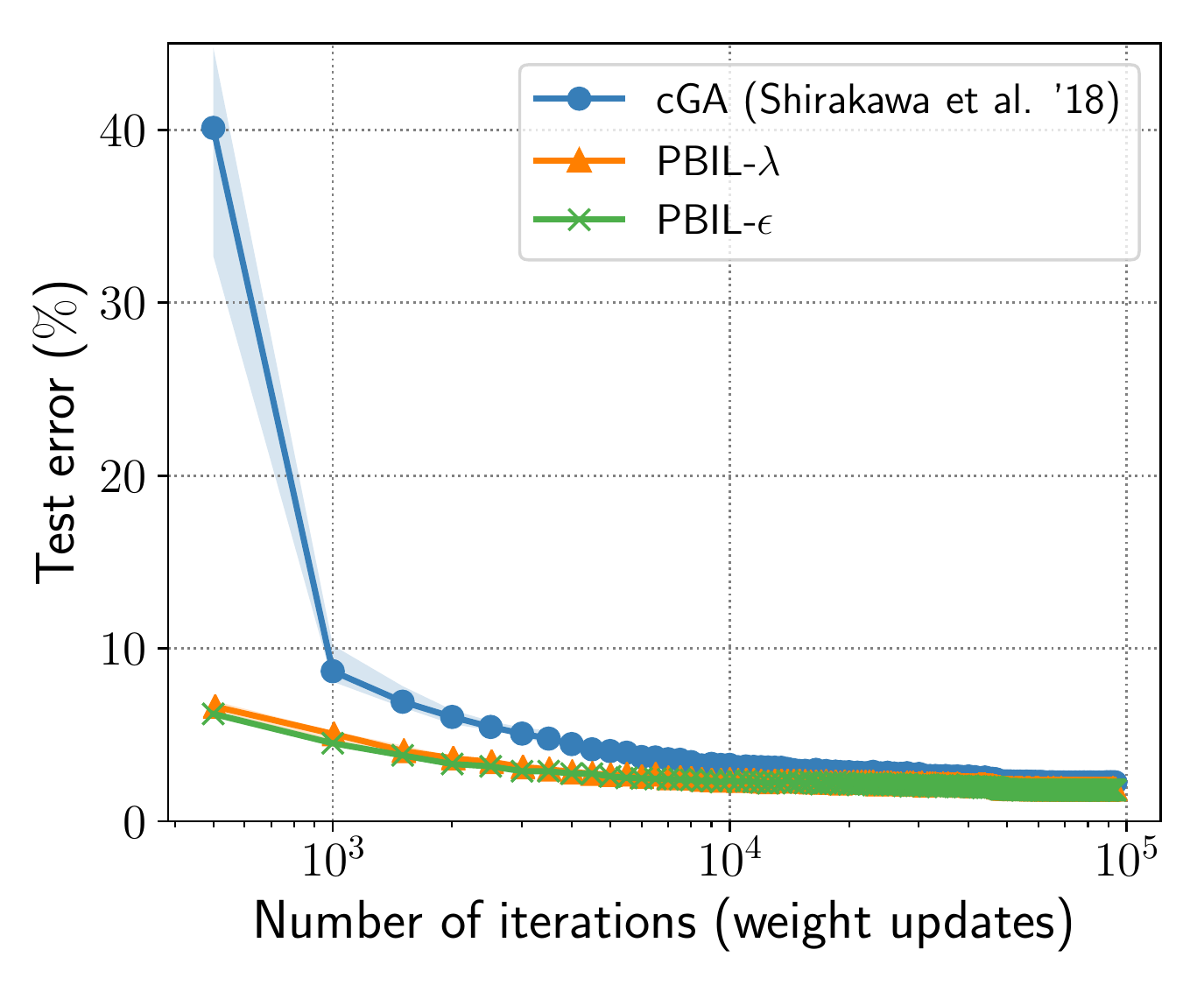}%
    \caption{Test error \new{(log-scale for the horizontal axis)}}%
    \label{fig:testerror}%
  \end{subfigure}%
  \begin{subfigure}{0.5\hsize}%
    \centering%
    \includegraphics[width=\hsize]{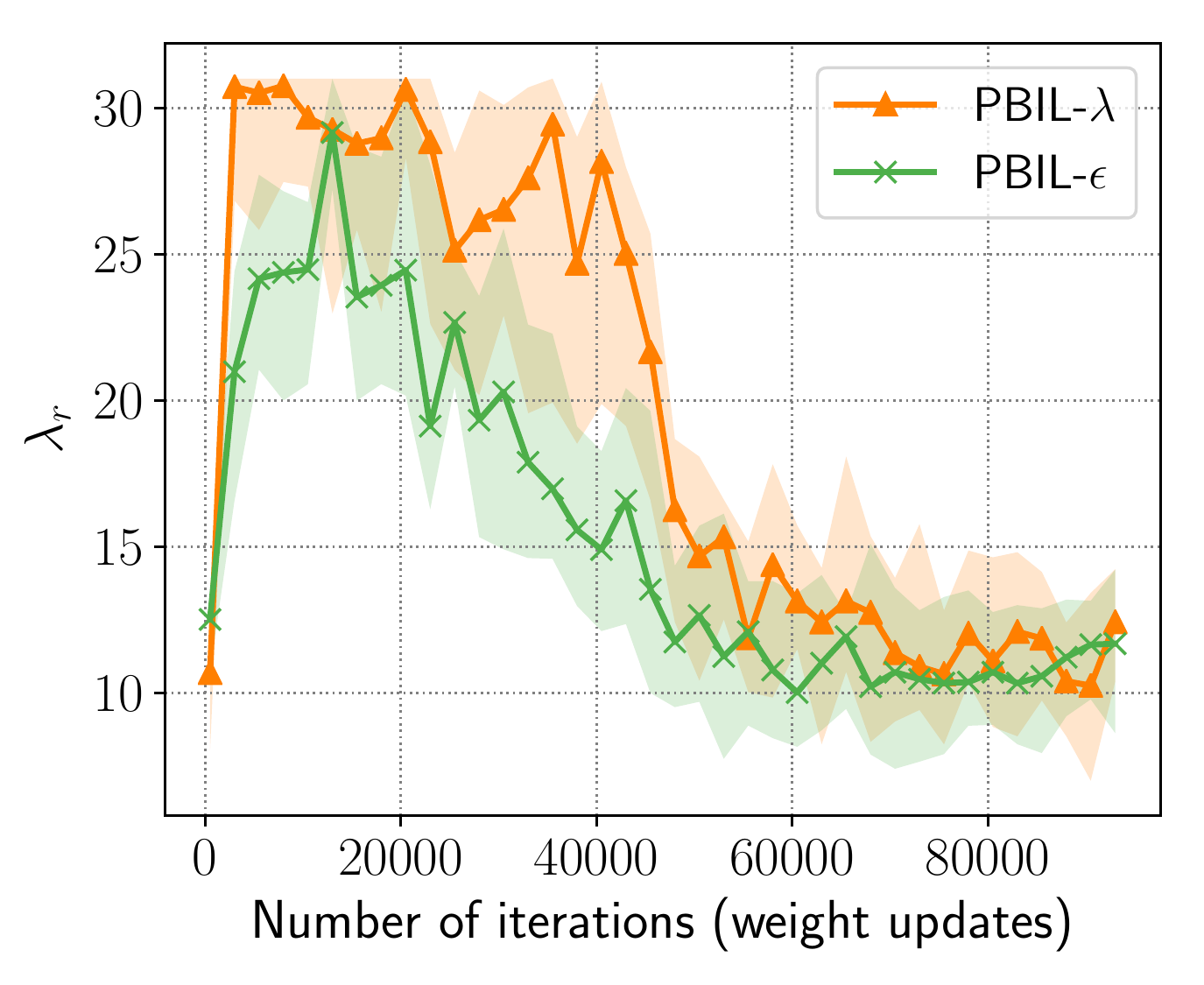}%
    \caption{Dynamics of $\lambda_r$\\ \ }%
    \label{fig:lam}%
  \end{subfigure}%
  \caption{Results on dynamic layer selection. Median (solid line) and inter-quartile range (transparent region).}%
  \label{fig:layers}%
\end{figure}

Figure~\ref{fig:layers} shows the test error over the number of weight updates (i.e., GPU processes) and the dynamics of $\lambda_r$ for \pbill\ and \pbile. The test errors of our methods outperform the previous cGA at not only the beginning stage of training but also the final iteration. The median (lower, upper quartiles of) test errors of cGA, \pbill\, and \pbile\ at the final iteration are $2.240$ ($2.130$, $2.298$), $1.855$ ($1.790$, $1.932$), and $1.795$ ($1.752$, $1.838$), respectively.

\paragraph{(II) Selection of Activation Functions}
A DNN with 3 full connected hidden layers with 1024 units for each layer was trained. A binary hyper-parameter vector $M \in \{0, 1\}^{n}$ of dimension $n = 3072$ encodes the type of activation for each unit: tanh ($\new{m}_i = 0$) or ReLU ($\new{m}_i = 1$).

\begin{figure}[t]
  \centering
  \begin{subfigure}{0.77\hsize}%
    \centering
    \includegraphics[width=\hsize]{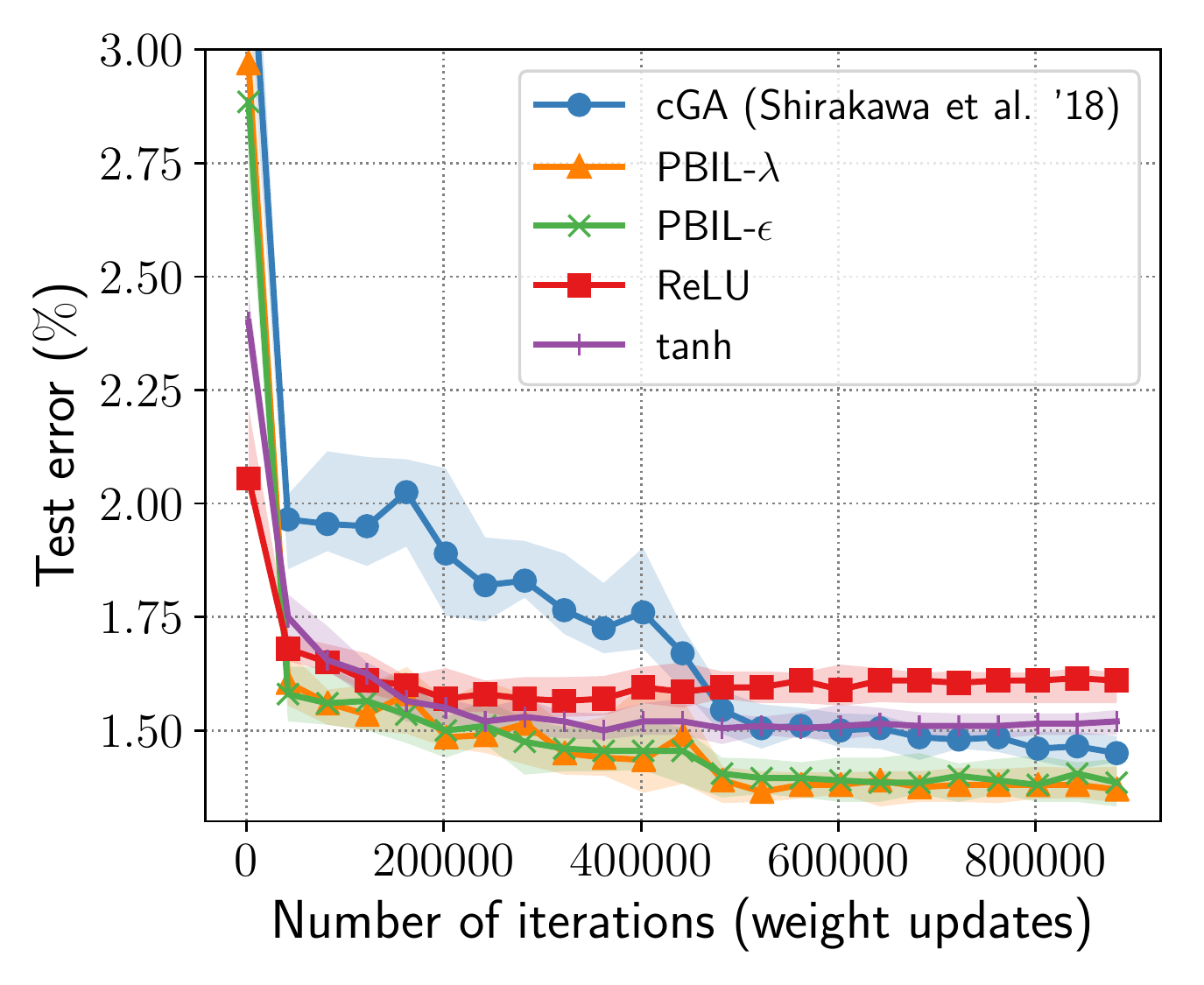}%
    \caption{Test error}%
    \label{fig:testerror}%
  \end{subfigure}%
  \\
  \begin{subfigure}{0.77\hsize}%
    \centering
    \includegraphics[width=\hsize]{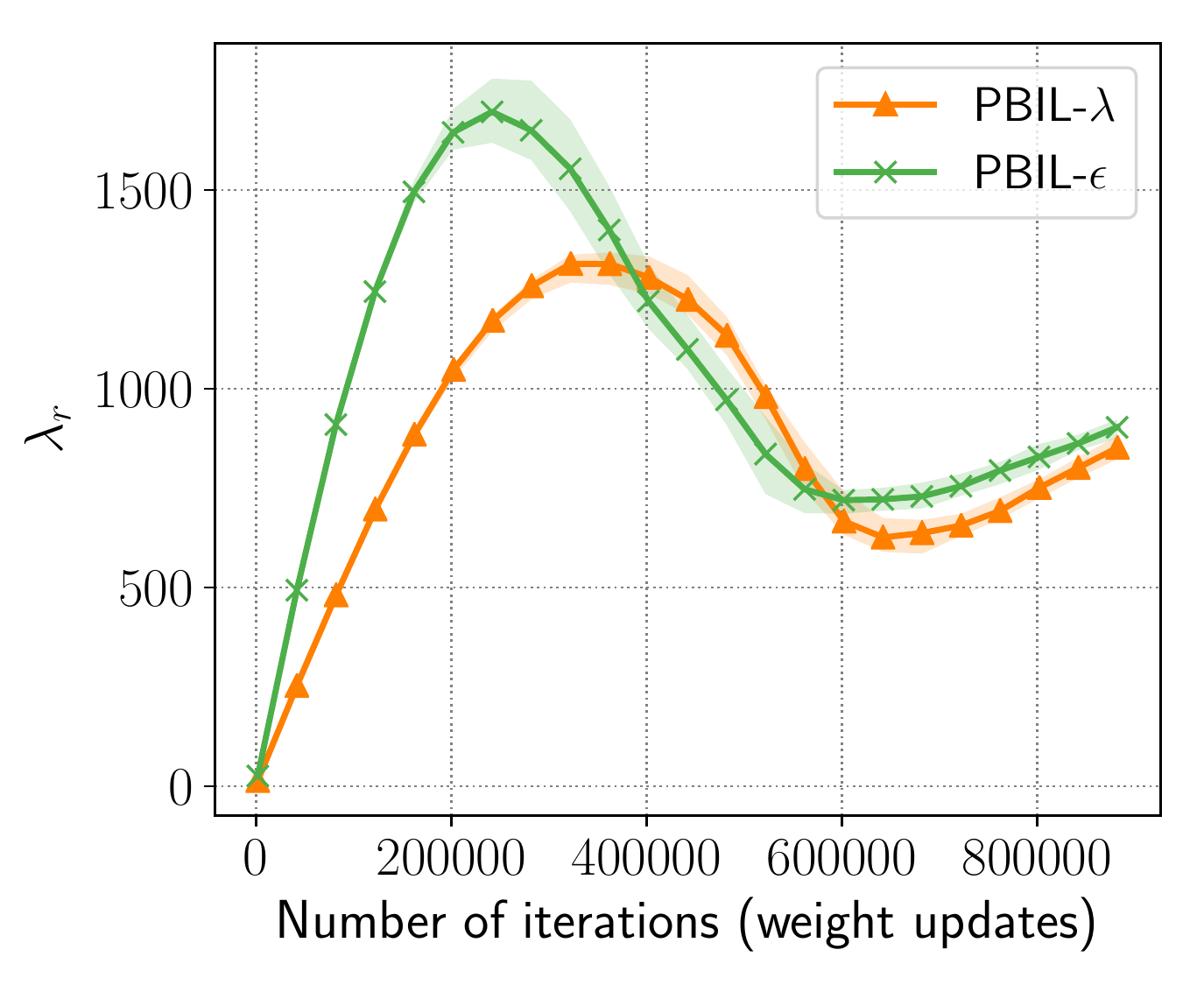}%
    \caption{Dynamics of $\lambda_r$}%
    \label{fig:lam}%
  \end{subfigure}%
  \caption{Results on dynamic activation selection. Median (solid line) and inter-quartile range (transparent region).
    The median (and lower, upper quartile) of the final test error of each algorithm are as follows;
    cGA: 1.450 (1.430, 1.488),
    \pbill: 1.370 (1.343, 1.420),
    \pbile: 1.385 (1.332, 1.437), 
    ReLU: 1.610 (1.560, 1.627),
    tanh: 1.520 (1.490, 1.545).
  }
  \label{fig:activations}
\end{figure}

Figure~\ref{fig:activations} shows the test error over the number of weight updates and the dynamics of $\lambda_r$ for \pbill\ and \pbile. The settings for cGA and the fixed $M$ cases (using ReLU or tanh for all hidden units) are the same as \cite{Shirakawa2018aaai}. The previous approach (cGA) had an advantage over fixed $M$ cases in the final performance, while the decrease of the test error was slower than the learning with fixed $M$. This disadvantage was mitigated by our proposed methods. Moreover, they outperformed cGA in the final performance. The step-size in \pbile\ and the sample size in \pbill\ were rather dynamic. In case of \pbile, $\epsilon = n^{-\frac12} / (\lambda_r / \lambda_{\min})$ became much smaller than that of cGA ($\epsilon = n^{-1}$), as the objective is regarded as a dynamic and noisy function. To conclude, the proposed approaches not only mitigated the strategy parameter tuning, but also enjoyed their dynamic change, improving both in speed and final performance. 



\section{Conclusion}

We developed a quasi-parameterless stochastic natural gradient based algorithm for black box discrete optimization. The algorithm is instantiated with Bernoulli distributions for binary optimization. The efficacy and the robustness of this approach was tested on simple yet descriptive and commonly used benchmark problems\del{, and on max 3-SAT benchmarks}{}. It improved over the most commonly used parameter setting without any effort of parameter tuning.

We introduced our mechanism into two examples of simultaneous optimization of network configurations and connection weights for a DL model, revealing that the proposed approach is advantageous both in speed and performance as it takes advantage of dynamic change of the strategy parameters. Again, we emphasize that the observed performance improvements were achieved without any strategy parameter tuning, which is the main point of this study.

This paper is a proof of concept, as we proposed a generic framework for optimization in arbitrary discrete search domain, while we have evaluated its single instantiation on binary search domain. Applications optimization on search space other than binary space such as optimization of categorical variables and mixed type variables are our important future work. Further applications to hyper-parameter optimization of DL configuration as well as engineering tasks such as power system controller design are also required to evaluate the usefulness of the proposed approach.

\section{Acknowledgement}
This work is partially supported by the SECOM Science and Technology Foundation.



\appendix
\section{Appendix: Benchmark Results}
\begin{figure}[h]
	\centering
	\includegraphics[width=\hsize]{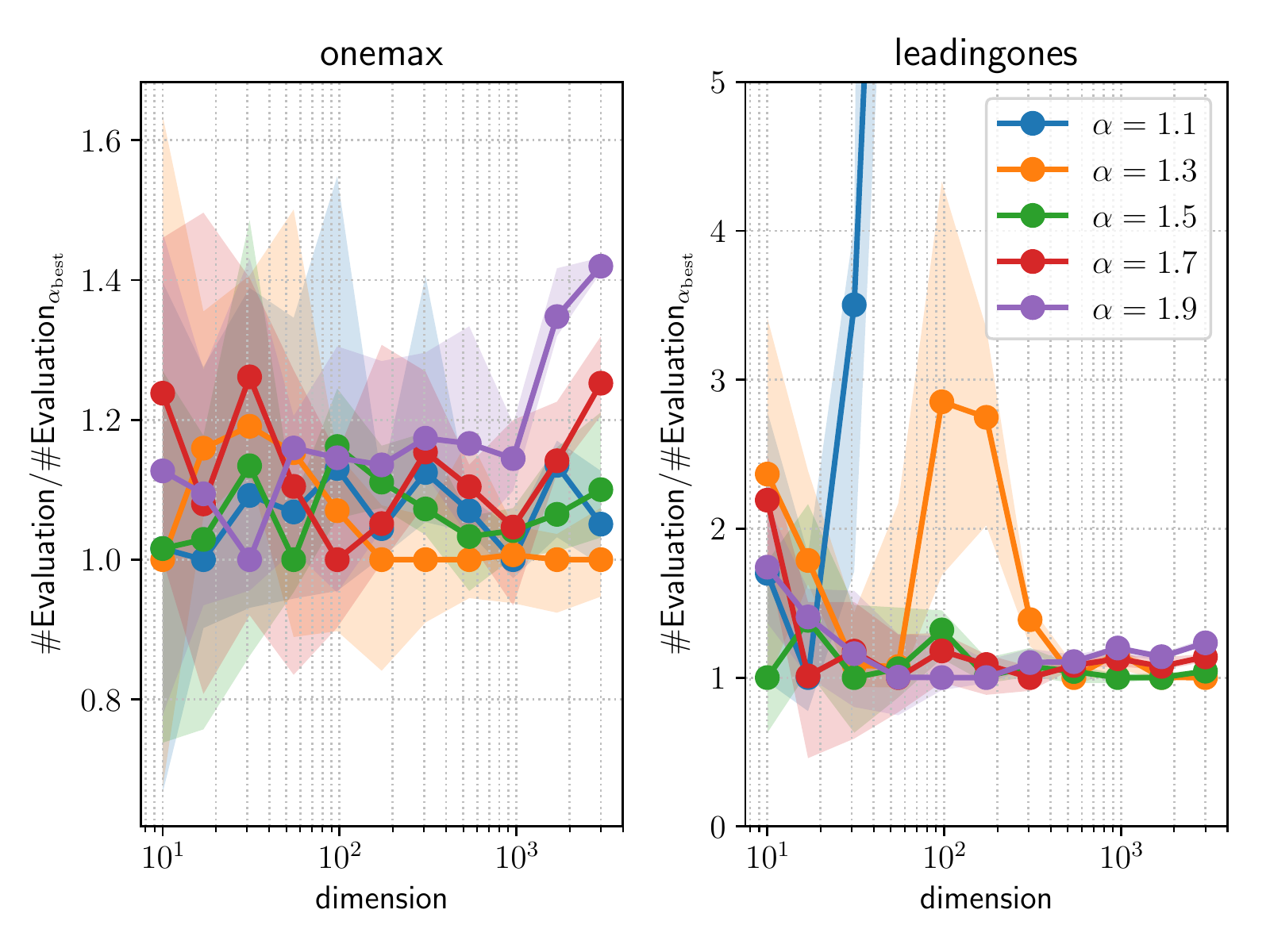}
	\caption{Robustness of \pbill\ against $\alpha$ variation. The median fractions over the best median value of the first hitting time are reported as well as its inter-quartile range. }
	\label{fig:sp2}
\end{figure}

We show an additional result to see the robustness of \pbill\ performance against $\alpha$ variation. 
Figure~\ref{fig:sp2} shows the performance differences of \pbill\ by different $\alpha$ values. The experimental setting is as written in the main text. The vertical axis indicates the first-hitting time divided by the median of the first-hitting time of the best $\alpha$ value in each dimension. On \textsc{OneMax}, even with the worst $\alpha$ value, the performance is at most 1.4 times worse compared to the best setting. On the other hand, $\alpha=1.1$ led to significant performance deterioration on $\textsc{LeadingOnes}$. However, larger values ($\alpha \geq 1.5$) resulted in stable performance as well.

\del{
\paragraph{Max 3 SAT Problems}

\pbill, \pbile\ and cGA with the default setting were compared on max 3-SAT benchmark problems.
We used uniform random-3-SAT instances taken from SATLIB\del{\footnote{http://www.cs.ubc.ca/~hoos/SATLIB/benchm.html}}{} with dimension (number of variables) 20, 50 and 75\del{ and 100\todo{keep updated}}{}. \todo{more information, e.g., number of clauses} The number of clauses depends on the dimension;  91, 218 and 325\del{ and 430}{} clauses exist on dimension 20, 50 and 75\del{ and 100}{}, respectively. We randomly picked out $10$ instances for each dimension. To turn 3-SAT to max 3-SAT problems, we defined the objective as the number of all clauses minus the number of satisfied clauses. The results are summarized in Figure~\ref{fig:sat}. \del{The results are similar to those on \textsc{OneMax}, in that the default cGA ($\epsilon = n^{-1}$) is too naive and cGA with $\epsilon = n^{-\frac12}$ performed faster. Both \pbill\ and \pbile\ performed better than the default cGA without any parameter tuning.}{}\new{\pbill\ and \pbile\ performed similarly, while significantly large variations were observed for cGA. For $n = 50$, the upper quartile of the number of $f$-calls reached $10^7 \times n$, indicating that more than $25\%$ of instances could not be solved within this budget. }

\begin{figure}[h]
  \centering%
  \includegraphics[width=0.8\hsize]{maxsat.pdf}%
  \caption{Results on max 3 SAT benchmarks.\todo{remove $\epsilon = 1/\sqrt{n}$, legend without parameters, x-axis in linear scale}}%
  \label{fig:sat}
\end{figure}
}{}

\section{Appendix: Hyper-Parameter Optimization}

We describe the procedure of the simultaneous optimization of binary hyper parameters and connection weights for DL models. Since the original cGA is equivalent to \pbile\ with $\lambda_{\min} = \lambda_{\max} = 2$ and $\epsilon = n^{-1}$, we describe the procedures using \pbill\ and \pbile. The procedures are summarized below. The number of $W$ update is denoted by $T$, and $T_{\max}$ is its maximum number. 

\paragraph{Simultaneous Optimization Using \pbile}
\begin{algorithmic}[1]
  \While{$\mathrm{T} < \mathrm{T}_{\max}$}
  \State draw a mini-batch $\mathcal{Z}$
  \State sample $M_1$ and $M_2$ independently from $P_\theta$
  \State compute losses $\mathcal{L}(W, M_i; \mathcal{Z})$ for $i=1,2$
  \State update $W$ using \eqref{eq:nggw}
  \State perform one step of \pbile\ (update $\theta$ and $\epsilon$)
  \State $\mathrm{T} \gets \mathrm{T} + 1$
  \EndWhile
\end{algorithmic}
As described in the main text, we update $W$ after every GPU process. Evaluation of two network configurations, $M_1$ and $M_2$, are done in one GPU process. 

\paragraph{Simultaneous Optimization Using \pbill}
\begin{algorithmic}[1]
  \While{$\mathrm{T} < \mathrm{T}_{\max}$}
  \For{$i = 1,\dots, \lambda$}
  \State draw a mini-batch $\mathcal{Z}_i$
  \State sample $M_i$ from $P_\theta$
  \State compute losses $\mathcal{L}(W, M_i; \mathcal{Z}_i)$
  \State update $W$ using $\nabla_W \mathcal{L}(W, M_i; \mathcal{Z}_i)$
  \State $\mathrm{T} \gets \mathrm{T} + 1$  
  \EndFor  
  \State perform one step of \pbill\ (update $\theta$ and $\lambda$)
  \EndWhile
\end{algorithmic}
For \pbill, different mini-batches are taken for different configurations $M_i$, and the losses are evaluated one-by-one. We have used the same mini-batch size $N$ for \pbill\ and \pbile\ to minimize the differences. However, since \pbile\ evaluates the losses for $M_1$ and $M_2$ in one GPU process, its computational road corresponds to passing a mini-batch of $2N$. It might be fair for \pbill\ to use mini-batch of size $2N$, which will further improve the performance of \pbill.

\end{document}